\documentclass[compact]{jetbrains_report}

\usepackage[style=authoryear,natbib=true,backend=biber]{biblatex}
\usepackage{pgfplots}
\usepackage{tikz}
\usetikzlibrary{positioning, arrows.meta, shapes.geometric, calc}
\pgfplotsset{compat=1.18}

\usepackage[capitalize,nameinlink]{cleveref}

\usepackage{fancyvrb}
\tcbuselibrary{breakable}

\usepackage{xspace}
\usepackage[nolist,nohyperlinks]{acronym}

\AtBeginDocument{%
  \makeatletter
  \def\@bsphack{%
    \relax\ifhmode
      \@savsk\lastskip
      \@savsf\spacefactor
      \ifnum\@savsf=\z@\@savsf=\@m\fi
      \spacefactor\z@
    \fi}%
  \def\@esphack{%
    \relax\ifhmode
      \spacefactor\@savsf
      \ifdim\@savsk>\z@
        \ifdim\lastskip=\z@
          \nobreak \hskip\z@skip
        \fi
        \ignorespaces
      \fi
    \fi}%
  \makeatother
}

\colorlet{GoodColor}{jb-teal!38}
\colorlet{MistakeColor}{jb-red!32}
\colorlet{UnnecessaryColor}{jb-yellow!60}

\hypersetup{colorlinks=true, linkcolor=jb-main, citecolor=jb-main, urlcolor=jb-main}

\xspaceaddexceptions{-}

\begin{acronym}
    \acro{llm}[LLM]{Large Language Model}
    \acro{rft}[RFT]{Rejection sampling Fine-Tuning}
    \acro{srft}[SRFT]{Step Rejection Fine-Tuning}
    \acro{her}[HER]{Hindsight Experience Replay}
\end{acronym}

\title{Step Rejection Fine-Tuning: A Practical Distillation Recipe}

\jbmetadata[s]{Date}{April 2026}

\jbauthor{Igor Slinko}{}
\jbauthor{Ilia Zavidnyi}{}
\jbauthor{Egor Bogomolov}{}
\jbauthor{Yaroslav Zharov}{}
\makeatletter
\g@addto@macro\jb@affiliations{JetBrains Research}
\makeatother
\jbcorresponding[{igor.slinko, ilia.zavidnyi, egor.bogomolov, yaroslav.zharov}@jetbrains.com]{igor.slinko@jetbrains.com}

\begin{document}

\maketitle

\begin{jbabstract}
\leavevmode\ac{rft} is a standard method for training \acs{llm} agents, where unsuccessful trajectories are
discarded from the training set. In the context of \texttt{SWE-bench} tasks, this corresponds to
filtering out runs where the submitted patch does not pass the tests. However, this approach
discards unresolved trajectories, even though they form a large portion of all trajectories for
hard tasks and even then may be partially correct. In this work, we propose
Step~Rejection~Fine-Tuning~(SRFT)\acused{srft}---a practical way to leverage these unresolved
trajectories. For this, we employ a critic \acs{llm} to assess the correctness of each step in a
trajectory. Consequently, during training, we mask the loss for erroneous steps while retaining
them in the context window. This way we ensure the model learns to recover from errors without
reproducing them. Evaluation on \texttt{SWE-bench~Verified} shows that while \ac{rft} improves
the resolution rate by $2.4\%$ by excluding unresolved trajectories, \ac{srft} improves it by
$3.7\%$ by filtering them instead of discarding completely, reaching the total resolution rate
of $32.2\%$.
\end{jbabstract}

\acresetall
\section{Introduction}

LLM-based agents are systems designed to autonomously perceive and interact with environments to
achieve complex goals~\citep{Xi2023,Wang2023}, with the ability to reason, plan, and use tools to
solve open-ended problems. LLM-based agents often employ paradigms like
ReAct~\citep{yao2023react} or Reflexion~\citep{Shinn2023} that interleave reasoning and acting
phases, constituting the so-called trajectory that consists of multiple such steps.

\leavevmode\ac{rft}~\citep{yuan2023scaling} has emerged as a standard paradigm for training LLM
agents~\citep{pan2025training,R2EGym2025,yang2025swesmith}. This process involves generating
multiple trajectories for a given task, filtering for those that successfully resolve the task,
and performing supervised fine-tuning on this filtered set. Filtering is needed, since
incorporating unresolved trajectories into the training set degrades performance by teaching the
model to imitate erroneous actions. A major limitation of \ac{rft} is the inability to leverage
unresolved trajectories. For instance, SWE-smith~\citep{yang2025swesmith} introduces a
large-scale dataset of agent trajectories obtained by solving synthesized software engineering
tasks and a model trained on this dataset. However, due to \ac{rft}, they discard approximately
$61\%$ of the collected runs, therefore losing a lot of potentially informative data.

Our key insight is that unresolved trajectories are not entirely erroneous; rather, they often
consist of correct and useful steps interspersed with errors. Our manual analysis of 20
trajectories indicates that even in unresolved trajectories, only up to $24\%$ of steps can be
classified as mistakes. To capitalize on this finding, we introduce \ac{srft}.
In this method, we utilize a critic \ac{llm} to discriminate trajectories on a lower level,
marking singular steps of trajectories as either worthy or unworthy to train on.
This method allows the model to learn from the valid portions of unresolved trajectories without
internalizing mistakes.

The contributions of this paper are as follows. We present a practical light-weight approach that
allows utilizing the data routinely considered noise without changing the overall pipeline.
The approach is detailed in \Cref{sec:method}. Then, we conduct an experiment that shows that on
the challenging \texttt{SWE-bench Verified} benchmark, our method successfully improves over the
naïve RFT baseline. The experiments are described and analyzed in \Cref{sec:experiment_setup}.

\section{Related Work}
\label{sec:related_work}

Research has increasingly moved beyond standard \acl{rft}---which treats all actions in a
resolved trajectory as equally valid---towards extracting granular supervision from suboptimal or
failed attempts.

\emph{Learn-by-interact}~\citep{LearnByInteract2501_10893}, an instance of a broader \ac{her}
approach~\citep{hindsight_experience_replay}, addresses instruction--trajectory misalignment in
long trajectories. They propose a backward construction method that decomposes trajectories into
shorter segments and synthesizes specific tasks for each, thereby creating valid demonstrations
for these synthesized tasks. While this and other \ac{her}-based methods effectively utilize
available data, they rely on generating synthetic instructions. In contrast, our approach focuses
on filtering steps within the original task context, avoiding the need for task synthesis. Also,
our method alleviates the risk of drifting off of the ground truth task.

\emph{SWEET-RL}~\citep{SWEETRL2503_15478} tackles the credit assignment problem by transforming
trajectory-level feedback into step-wise signals. It trains a step-wise critic by comparing pairs
of trajectories and constraining the score to be a sum of per-step contributions, forcing an
implicit decomposition of preferences. This critic then guides policy optimization via Direct
Preference Optimization~\citep[DPO;][]{rafailov2023direct}. While effective, this introduces the
complexity of Reinforcement Learning. In contrast, our method avoids training a critic with
complex objectives, offering a simpler alternative within the supervised learning paradigm.

Finally, \emph{STeP}~\citep{SelfRefPartialMasking2505_20023} employs a teacher-in-the-loop to
actively synthesize ``self-reflected'' trajectories where errors are immediately followed by
teacher-generated reflections and corrections. By applying partial masking to error steps, they
enable the model to learn recovery strategies. While our work shares the core mechanism of
partial masking, we propose a simple practical approach that can retroactively fit into existing
pipelines. Specifically, instead of requiring expensive real-time teacher intervention to
synthesize new correction steps, we use an offline critic to salvage valid signals from standard
unresolved trajectories.

\section{Method}
\label{sec:method}

\subsection{Preliminaries}

We consider the problem of distilling the agentic behavior from a strong Teacher model to a
weaker Student model. The distillation process consists of gathering the outputs of the Teacher
and training the Student to mimic them. We denote the set of the collected Teacher outputs as
$\mathcal{D}$. We denote the Student model as a function $f_{\theta}$ parametrized by weights
$\theta$, that estimates the probability of the next token given the previous tokens.

For agentic behavior in particular, we need to sample coherent ReAct trajectories rather than
one-shot predictions, which complicates the dataset collection. Each trajectory
$\tau_i \in \mathcal{D}$ is defined as
$\tau_i = (s, u_i, (a_{i,0}, o_{i,0}), (a_{i,1}, o_{i,1}), \dots, (a_{i,T_i}, o_{i,T_i}))$,
comprising a system message~$s$, a task description~$u_i$, and a sequence of assistant actions
and corresponding environment observations $\{(a_t, o_t)\}_{t=0}^{T_i}$, where $T_i$ is the
length of the $i$-th trajectory in steps. For the sake of simplicity, we use
$\tau_{i,[0:t]}$ to denote the concatenation of the trajectory $\tau_i$ up to but not including
the $t$-th step, and $a_{t,[0:j]}$ to denote the string representation of the action $a_t$ up
to but not including $j$-th token.

\subsection{Methods Investigated in Current Work}

The \textbf{naïve distillation} is performed by minimizing the following loss
$\mathcal{L}(\theta,\mathcal{D})$, where NLL is the negative log-likelihood loss.
\begin{equation*} \label{eq:distill_loss}
    \mathcal{L}(\theta, \mathcal{D}) \;=\; \sum_{\tau_i\in\mathcal{D}} \sum_{t=0}^{T_i}
    \sum_{j=0}^{\vert a_t \vert}
    \text{NLL}\!\big(f_\theta(\tau_{i,[0:t]}+a_{t,[0:j]}), a_{t,j}\big)
\end{equation*}

The \textbf{\ac{rft}} filters the raw teacher trajectories in $\mathcal{D}$ to retain only
trajectories satisfying the final success criteria of the environment, resulting in the subset
$\mathcal{D}_s$. For example, in \texttt{SWE-bench}, it means that after the agent has finished,
a pre-defined set of tests should succeed. The Student model is then trained by minimizing the
loss $\mathcal{L}(\theta,\mathcal{D}_s)$. However, this approach discards the set of unresolved
trajectories $\mathcal{D}_{f} = \mathcal{D} \setminus \mathcal{D}_{s}$, which may be large for
complex tasks.

To leverage the data rejected under the \ac{rft} framework, we introduce a fine-grained
supervision mechanism \textbf{\ac{srft}}. Instead of discarding or keeping the entire
trajectories, we introduce weights $\mathcal{W}$ to keep alongside the trajectories. Weight
$w_{i,t} \in \mathcal{W}$ corresponds to the step $(a_{i,t}, o_{i,t})$ in trajectory $\tau_i$.
Given the weights, we modify the loss to perform the weighted distillation on the whole dataset
with loss $\mathcal{L}(\theta, \mathcal{D}, \mathcal{W})$.
\begin{equation*} \label{eq:distill_loss_weighted}
    \mathcal{L}_\mathcal{W}(\theta, \mathcal{D}, \mathcal{W}) \;=\;
    \sum_{\tau_i\in\mathcal{D}'}\sum_{t=0}^{T_i} w_{i,t} \cdot
    \sum_{j=0}^{\vert a_t \vert}
    \text{NLL}\!\big(f_\theta(\tau_{i,[0:t]}+a_{t,[0:j]}), a_{t,j}\big)
\end{equation*}

\subsection{\ac{srft} Instantiation in Current Work}

\ac{srft} supports different weighting schemes, but for our current experiments, we instantiate
it as follows. For successful trajectories in $\mathcal{D}_s$, we assign a weight of $1$ to each
step. For unresolved trajectories in $\mathcal{D}_f$, we employ a critic model to label each
action $a_t$. We classify a step as \emph{good} if it advances the task, \emph{harmful} if it
hinders progress (e.g., introduces a bug), and \emph{unnecessary} if it neither helps nor
damages. We group \emph{good} and \emph{unnecessary} steps into a single category of
\emph{Productive} steps. Accordingly, we assign $w_t=0$ to harmful steps and $w_t=1$ otherwise,
forming $\mathcal{W}_f$. The total set of weights is $\mathcal{W} = \mathcal{W}_f \cup
\mathcal{W}_s$.

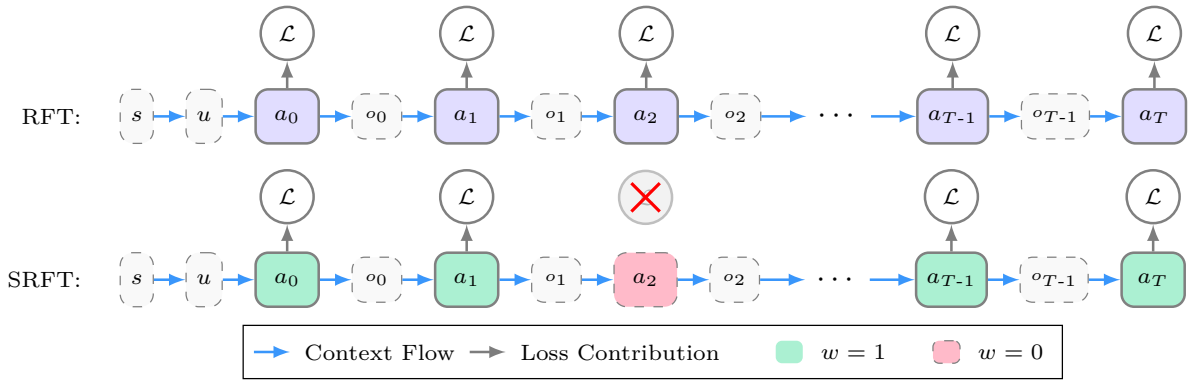
\begin{figure}[ht]
\centering
\resizebox{\textwidth}{!}{
\begin{tikzpicture}[
    node distance=0.25cm,
    prompt/.style={rectangle, draw=black!50, dashed, rounded corners, minimum height=0.6cm, align=center, font=\scriptsize, fill=gray!5},
    action/.style={rectangle, draw=black!50, thick, rounded corners, minimum width=0.7cm, minimum height=0.6cm, align=center, font=\scriptsize},
    obs/.style={rectangle, draw=black!50, dashed, rounded corners, minimum width=0.55cm, minimum height=0.5cm, align=center, font=\tiny, fill=gray!5},
    loss/.style={circle, draw=black!50, thick, minimum size=0.5cm, font=\scriptsize},
    context/.style={->, >=latex, thick, jb-main!80},
    contribution/.style={->, >=latex, thick, jb-midgray}
]

\node[prompt] (s1) {$s$};
\node[prompt, right=0.35cm of s1] (u1) {$u$};

\node[action, fill=jb-violet!20, right=0.35cm of u1] (a01) {$a_0$};
\node[obs, right=0.35cm of a01] (o01) {$o_0$};

\node[action, fill=jb-violet!20, right=0.35cm of o01] (a11) {$a_1$};
\node[obs, right=0.35cm of a11] (o11) {$o_1$};

\node[action, fill=jb-violet!20, right=0.35cm of o11] (a21) {$a_2$};
\node[obs, right=0.35cm of a21] (o21) {$o_2$};

\node[right=0.5cm of o21, font=\normalsize] (dots1) {$\cdots$};

\node[action, fill=jb-violet!20, right=0.5cm of dots1] (aT11) {$a_{T\text{-}1}$};
\node[obs, right=0.35cm of aT11] (oT11) {$o_{T\text{-}1}$};

\node[action, fill=jb-violet!20, right=0.35cm of oT11] (aT1) {$a_T$};

\draw[context] (s1) -- (u1);
\draw[context] (u1) -- (a01);
\draw[context] (a01) -- (o01);
\draw[context] (o01) -- (a11);
\draw[context] (a11) -- (o11);
\draw[context] (o11) -- (a21);
\draw[context] (a21) -- (o21);
\draw[context] (o21) -- (dots1);
\draw[context] (dots1) -- (aT11);
\draw[context] (aT11) -- (oT11);
\draw[context] (oT11) -- (aT1);

\node[loss, above=0.3cm of a01] (l01) {$\mathcal{L}$};
\draw[contribution] (a01) -- (l01);

\node[loss, above=0.3cm of a11] (l11) {$\mathcal{L}$};
\draw[contribution] (a11) -- (l11);

\node[loss, above=0.3cm of a21] (l21) {$\mathcal{L}$};
\draw[contribution] (a21) -- (l21);

\node[loss, above=0.3cm of aT11] (lT11) {$\mathcal{L}$};
\draw[contribution] (aT11) -- (lT11);

\node[loss, above=0.3cm of aT1] (lT1) {$\mathcal{L}$};
\draw[contribution] (aT1) -- (lT1);

\node[font=\scriptsize, anchor=east] at ([xshift=-0.3cm]s1.west) {RFT:};

\node[prompt, below=1.2cm of s1] (s) {$s$};
\node[prompt, right=0.35cm of s] (u) {$u$};

\node[action, fill=GoodColor, right=0.35cm of u] (a0) {$a_0$};
\node[obs, right=0.35cm of a0] (o0) {$o_0$};

\node[action, fill=GoodColor, right=0.35cm of o0] (a1) {$a_1$};
\node[obs, right=0.35cm of a1] (o1) {$o_1$};

\node[rectangle, draw=black!50, dashed, rounded corners, minimum width=0.7cm, minimum height=0.6cm, align=center, font=\scriptsize, fill=MistakeColor, right=0.35cm of o1] (a2) {$a_2$};
\node[obs, right=0.35cm of a2] (o2) {$o_2$};

\node[right=0.5cm of o2, font=\normalsize] (dots) {$\cdots$};

\node[action, fill=GoodColor, right=0.5cm of dots] (aT1_) {$a_{T\text{-}1}$};
\node[obs, right=0.35cm of aT1_] (oT1) {$o_{T\text{-}1}$};

\node[action, fill=GoodColor, right=0.35cm of oT1] (aT) {$a_T$};

\draw[context] (s) -- (u);
\draw[context] (u) -- (a0);
\draw[context] (a0) -- (o0);
\draw[context] (o0) -- (a1);
\draw[context] (a1) -- (o1);
\draw[context] (o1) -- (a2);
\draw[context] (a2) -- (o2);
\draw[context] (o2) -- (dots);
\draw[context] (dots) -- (aT1_);
\draw[context] (aT1_) -- (oT1);
\draw[context] (oT1) -- (aT);

\node[loss, above=0.3cm of a0] (l0) {$\mathcal{L}$};
\draw[contribution] (a0) -- (l0);

\node[loss, above=0.3cm of a1] (l1) {$\mathcal{L}$};
\draw[contribution] (a1) -- (l1);

\node[circle, draw=black!25, thick, minimum size=0.5cm, font=\scriptsize, fill=gray!10, above=0.3cm of a2] (l2) {\textcolor{black!30}{$\mathcal{L}$}};
\draw[red, thick, line width=1pt] ($(l2.north west)+(0.06,-0.06)$) -- ($(l2.south east)+(-0.06,0.06)$);
\draw[red, thick, line width=1pt] ($(l2.north east)+(-0.06,-0.06)$) -- ($(l2.south west)+(0.06,0.06)$);

\node[loss, above=0.3cm of aT1_] (lT1_) {$\mathcal{L}$};
\draw[contribution] (aT1_) -- (lT1_);

\node[loss, above=0.3cm of aT] (lT) {$\mathcal{L}$};
\draw[contribution] (aT) -- (lT);

\node[font=\scriptsize, anchor=east] at ([xshift=-0.3cm]s.west) {SRFT:};

\coordinate (center) at ($(s.west)!0.5!(aT.east)$);
\node[draw, font=\scriptsize, inner sep=3pt, anchor=north] (legend) at ($(center)+(0,-0.5)$) {%
  \begin{tikzpicture}[baseline=-0.5ex]
    \node[font=\scriptsize] (cf) at (0,0) {Context Flow};
    \draw[->, >=latex, thick, jb-main!80] ([xshift=-0.45cm]cf.west) -- ([xshift=-0.05cm]cf.west);
    \node[font=\scriptsize, right=0.5cm of cf] (lc) {Loss Contribution};
    \draw[->, >=latex, thick, jb-midgray] ([xshift=-0.45cm]lc.west) -- ([xshift=-0.05cm]lc.west);
    \node[rectangle, fill=GoodColor, rounded corners=2pt, minimum width=0.3cm, minimum height=0.25cm, right=0.5cm of lc] (gc) {};
    \node[font=\scriptsize, right=0.08cm of gc] (gt) {$w=1$};
    \node[rectangle, fill=MistakeColor, rounded corners=2pt, draw=black!50, dashed, minimum width=0.3cm, minimum height=0.25cm, right=0.4cm of gt] (bc) {};
    \node[font=\scriptsize, right=0.08cm of bc] {$w=0$};
  \end{tikzpicture}%
};

\end{tikzpicture}
}
\caption{Comparison of RFT and SRFT training approaches.}
\label{fig:trajectory_example}
\end{figure}

Given this weighting scheme, the loss $\mathcal{L}_\mathcal{W}$ simply omits the loss
calculation for the steps that were marked as harmful. Conveniently, for binary weights, this
corresponds to modifying token-wise masks. This is routinely done for observations, and including
masks for some pre-defined steps incurs little engineering overhead. A further example of how
this is applied to a trajectory is depicted in~\Cref{fig:trajectory_example}.

\FloatBarrier

\section{Experimental Setup}
\label{sec:experiment_setup}

To experimentally demonstrate that \ac{srft} improves the performance of the Student model, we
employ it on a complex Software Engineering task. We fine-tune
\texttt{Qwen2.5-Coder-32B-Instruct}~\citep{qwen2024coder} as our Student model, using the
SWE-Agent~\citep{SWEAgent2024} framework as the scaffold. For training, we utilize the
\texttt{SWE-smith-trajectories} dataset~\citep{yang2025swesmith}, which consists of approximately
25,000 trajectories generated by \texttt{SWE-agent}. The dataset comprises $39\%$ resolved and
$61\%$ unresolved trajectories. For our study, we sample a balanced set of 5,000 resolved
($\mathcal{D}_s$) and 5,000 unresolved ($\mathcal{D}_f$) trajectories. We further instantiate
datasets for different methods as described in~\Cref{sec:method}.

To provide the markup for the \ac{srft} method, we employ Claude 4 Sonnet (snapshot
20250514)~\citep{anthropic2025claude4} as the critic model, incurring a total cost of \$$660$
for 5,000 trajectories. The critic prompt was selected by expert labeling of trajectories, and is
presented alongside the expert data in~\Cref{app:critic_prompt}. \Cref{tab:label_stats_table}
details the distribution of step labels. We note that unresolved trajectories contain more steps
marked as harmful, but the critic was never provided with the resolution status of the
trajectory. This aligns with the intuition that the unresolved trajectories contain potentially
adversarial patterns not to be distilled.

\begin{table}[t]
  \centering
  \begin{tabular}{lcc}
  \toprule
  Label & Resolved & Unresolved \\
  \midrule
  Productive steps & 95.9\% & 93.0\% \\
  Harmful steps & \phantom{0}4.1\% & \phantom{0}7.1\% \\
  \bottomrule
\end{tabular}
  \caption{Label distribution for 5,000 resolved and 5,000 unresolved trajectories.}
  \label{tab:label_stats_table}
\end{table}

We evaluate the performance of the Student model on the \texttt{SWE-bench Verified}
dataset~\citep{SWEbenchVerifiedOpenAI2024}. To ensure robustness, each experiment is repeated 7
times. We report the Resolved Rate in percent. The results are presented in
\Cref{tab:main}.

Consistent with our hypothesis, naively including unresolved trajectories leads to a performance
degradation compared to the RFT baseline (28.5\% vs 30.9\%), as the model internalizes errors
present in the failed attempts. However, by applying critic-guided masking, we not only mitigate
this degradation but achieve a performance gain, outperforming the RFT baseline (32.2\%
vs 30.9\%). This improvement is statistically significant; a bootstrap analysis confirms a gain
of 1.3\% with a 95\% confidence interval of [0.4, 2.3] (refer to
\Cref{app:significance} for detailed statistical analysis).

\newcommand{\pdiff}[1]{\,\raisebox{1pt}{\textcolor{jb-teal}{\scriptsize(+#1)}}}
\newcommand{\ndiff}[1]{\,\raisebox{1pt}{\textcolor{jb-red}{\scriptsize(-#1)}}}

\begin{table}[t]
  \centering
  \begin{tabular}{ll}
  \toprule
  Training Data & Resolved (\%)\\
  \midrule
  Base model & \phantom{0}7.0 $\pm$ 1.3 \ndiff{21.5}\\
  Naïve distillation & 28.5 $\pm$ 1.7 \\
  \ac{rft} & 30.9 $\pm$ 1.1 \pdiff{2.4} \\
  \ac{srft} & 32.2 $\pm$ 0.9 \pdiff{3.7}\\
  \bottomrule
  \end{tabular}
  \caption{Main results on \texttt{SWE-bench Verified}. Each experiment was run 7 times; we
  report mean $\pm$ standard deviation. Base model performance is from the paper
  R2E-Gym~\citep{R2EGym2025}, which uses the same scaffold and model.}
  \label{tab:main}
\end{table}

\section{Limitations and Future Work}

Our approach relies on the accuracy of the critic. Mislabeling valid steps as harmful can reduce
the effective training data, while failing to identify subtle errors can allow them to propagate
into the student model. We leave a thorough study of different critics and labeling approaches to
future work.

In this work, we did not explore the possibility of the loss weights
beyond binary. However, we note that this may be the key to further improving the quality, both
on resolved and unresolved trajectories.

While we test \ac{srft} on a challenging SWE-bench task, we acknowledge the lack of a
generalization study and leave the investigation of the method's generalizability to future work.

\section{Conclusion}

We have presented \acl{srft}, a straightforward yet effective enhancement to the \acl{rft}
distillation method that unlocks the value of unresolved trajectories. By selectively masking
steps, we enable agents to learn from the partial successes of the Teacher model within failed
attempts without internalizing its errors. On \texttt{SWE-bench Verified}, our method yields
statistically significant improvements over the RFT (32.2\% vs 30.9\% resolved issues),
highlighting the potential of step-level supervision in agentic distillation.

\FloatBarrier
\printbibliography

\newpage
\appendix

\section{Critic Prompt and Evaluation}
\label{app:critic_prompt}

We validated the critic prompt by manually labeling a ground truth set of 20 trajectories
(comprising 444 steps in total). \Cref{tab:critic_eval} details the performance of the
critic against these manual labels. The critic demonstrates robust performance in identifying
\emph{good} steps (F1=0.737), which constitutes the majority class. While performance on
minority classes (\emph{harmful} and \emph{unnecessary}) is lower, the overall system benefit
suggests that even imperfect filtering is preferable to naïve data inclusion or complete data
exclusion.

\begin{table}[h!]
  \centering
  \begin{tabular}{lccc}
  \toprule
  Label & F1 & Precision & Recall \\
  \midrule
  good & 0.737 & 0.660 & 0.835 \\
  unnecessary & 0.236 & 0.253 & 0.221 \\
  harmful & 0.384 & 0.604 & 0.282 \\
  \midrule
  Accuracy & \multicolumn{3}{c}{0.586} \\
  \bottomrule
  \end{tabular}
  \caption{Critic prompt evaluation on 444 manually labeled steps.}
  \label{tab:critic_eval}
\end{table}

The specific prompt used for the critic is provided below:

\begin{tcolorbox}[
  colback=gray!5,
  colframe=gray!50,
  boxrule=0.5pt,
  arc=2pt,
  fontupper=\small\ttfamily,
  breakable
]
You are evaluating SWE-agent trajectories.\\
Your goal is to annotate each step of the trajectory.\\
This annotation will be used to train a model only on good steps.\\
First, construct a brief "ideal solution path" and patch (in your own reasoning, not in the output).\\
Then, for each step, decide if it matches or deviates from this ideal path.\\
\\
\#\# Mark Each Step\\
- **"good"**: Part of or necessary for the ideal shortest path (exploration, correct fixes).\\
- **"harmful"**: Moves away from the path (wrong edits, irrelevant files), harmful tool usage, etc.\\
- **"unnecessary"**: Not harmful but not part of the path (redundant exploration, repeated actions).\\
\\
\#\# Rules\\
- Always compare each step to the shortest correct path you imagined.\\
- A step is "good" only if it is part of or necessary for that shortest path.\\
- A step is "harmful" if it deviates from or worsens the trajectory relative to that path.\\
- A step is "unnecessary" if it neither helps nor harms but is outside the shortest path.
\end{tcolorbox}

\clearpage
\section{Statistical Significance and Extended Results}
\label{app:significance}

Given the inherent high variance in LLM agent evaluation, we conducted a rigorous statistical
analysis. We identified two primary sources of noise: rollout variance, where two distinct
rollouts of the same model can differ by up to 4.8\%, and training variance, where two models
trained with different seeds can differ by 1.1\% on average across 7 rollouts.

\begin{figure}[ht]
\centering
\includegraphics[width=0.9\linewidth]{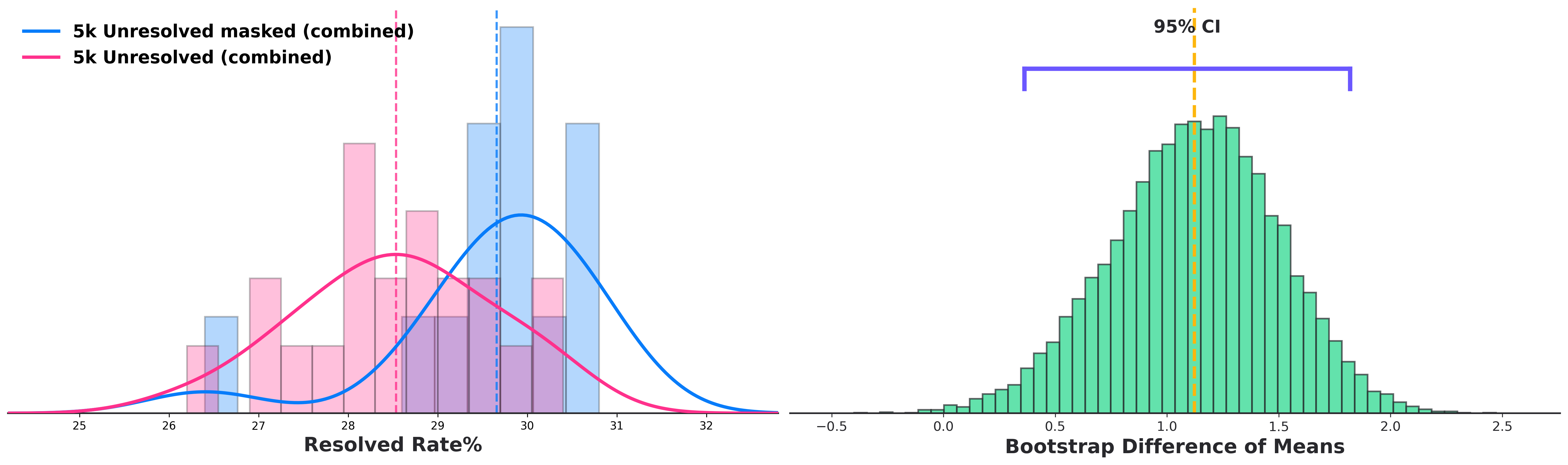}
\caption{Bootstrap analysis of Resolved Rate\% for 5k Unresolved vs 5k Unresolved masked.}
\label{fig:significance}
\end{figure}

To verify that our method provides a statistically significant improvement, we focused on the
unresolved split, which most clearly highlights the difference between using all steps versus
masking. We performed 5 training runs: 3 without masking and 2 with masking. Accounting for both
sources of noise, we still observe a performance increase (see \Cref{tab:significance}).
As shown in \Cref{fig:significance}, the bootstrap analysis confirms a statistically
significant improvement of 1.1\% with a 95\% confidence interval of [0.4, 1.8].

\begin{table}[h!]
  \centering
  \small
  \begin{tabular}{lcc}
  \toprule
  Configuration & Resolved\% & pass@7\% \\
  \midrule
  5k Unresolved (train \#1) & 28.8 $\pm$ 1.0 & 42.8 \\
  5k Unresolved (train \#2) & 27.7 $\pm$ 0.9 & 42.2 \\
  5k Unresolved (train \#3) & 29.1 $\pm$ 1.0 & 41.8 \\
  \rowcolor{gray!10} 5k Unresolved (combined) & 28.5 $\pm$ 1.1 & 42.3 \\
  \midrule
  5k Unresolved masked (train \#1) & 29.8 $\pm$ 0.6 & 42.6 \\
  5k Unresolved masked (train \#2) & 29.5 $\pm$ 1.5 & 43.8 \\
  \rowcolor{gray!10} 5k Unresolved masked (combined) & 29.7 $\pm$ 1.1 & 43.2 \\
  \bottomrule
  \end{tabular}
  \caption{Results from multiple training runs on 5k Unresolved trajectories. Mean and standard
  deviation are calculated over 7 rollouts for individual runs; combined results are calculated
  over 21 and 14 rollouts respectively.}
  \label{tab:significance}
\end{table}

\begin{table}[h!]
  \centering
  \small
  \begin{tabular}{llcc}
  \toprule
   & Training Data & Resolved\% & pass@7\% \\
  \midrule
  Base model & & \phantom{0}7.0 $\pm$ 1.3 & \\
  \midrule
  \ac{rft} & 5k Resolved & \underline{30.9} $\pm$ 1.1 & 45.8 \\
   & 5k Unresolved & 27.7 $\pm$ 0.9 & 42.2 \\
  Naïve distillation & 5k Resolved + 5k Unresolved & 28.5 $\pm$ 1.7 & 43.8 \\
  \ac{srft} & 5k Resolved + 5k Unresolved (masked) & \textbf{32.2} $\pm$ 0.9 & 45.8 \\
   & 5k Resolved (masked) + 5k Unresolved (masked) & 29.4 $\pm$ 1.6 & 41.2 \\
  \bottomrule
  \end{tabular}
  \caption{Extended experimental results on \texttt{SWE-bench Verified}.}
  \label{tab:full_results}
\end{table}

\end{document}